\documentclass[11pt]{article}
\usepackage{paclic32}
\usepackage{times}
\usepackage{latexsym}
\usepackage{amsmath}
\usepackage{multirow}
\usepackage{url}
\usepackage{booktabs}
\usepackage{adjustbox}
\usepackage{pbox}
\usepackage{array}
\usepackage{amsmath}
\usepackage{amssymb}
\usepackage{stmaryrd}
\usepackage{amsfonts}
\usepackage{subfigure}
 \usepackage{footmisc}
\usepackage[misc]{ifsym}
\usepackage[colorlinks,
           linkcolor=black,
            anchorcolor=blue,
            citecolor=blue
            ]{hyperref}
\usepackage[final]{changes}

\title{DEMN: Distilled-Exposition Enhanced Matching Network  \\for Story Comprehension}

\author{Chunhua Liu\textsuperscript{1}\quad  Haiou Zhang\textsuperscript{1}\quad  Shan Jiang\textsuperscript{1}\quad Dong Yu\textsuperscript{1,2} \Letter \\
 \textsuperscript{1} Beijing Language and Culture University \\
 \textsuperscript{2} Beijing Advanced Innovation for Language Resources of BLCU \\
 {\tt \{chunhualiu596, jiangshan727\}@gmail.com }\\
 {\tt \{hozhangel, yudong\_blcu\}@126.com  }\\}
\date{}

\begin{document}
\setlength{\abovedisplayskip}{4pt }      
\setlength{\belowdisplayskip}{6pt }
\maketitle
\begin{abstract} 
This paper proposes a Distilled-Exposition Enhanced Matching Network (DEMN) for story-cloze test, which is still a challenging task in story comprehension. We divide a complete story into three narrative segments: an \textit{exposition}, a \textit{climax}, and an \textit{ending}. The model consists of three modules: input module, matching module, and distillation module. The input module provides semantic representations for the three segments and then feeds them into the other two modules. The matching module collects interaction features between the ending and the climax. The distillation module distills the crucial semantic information in the exposition and infuses it into the matching module in two different ways. We evaluate our single and ensemble model on ROCStories Corpus \cite{Mostafazadeh2016ACA}, achieving an accuracy of 80.1\% and 81.2\% on the test set respectively. The experimental results demonstrate that our DEMN model achieves a state-of-the-art performance.



\end{abstract}

\section{Introduction}


 
Story comprehension is a fascinating task in natural language understanding with a long history \cite{Jones1974UnderstandingNL,turner1994creative}. The difficulty of this task arises from the necessity of commonsense knowledge, cross-sentence reasoning, and causal reasoning between events. 
The recently emerged story-cloze test \cite{Mostafazadeh2016ACA} focuses on commonsense story comprehension, which aims at choosing the most plausible ending from two options for a four-sentence story (also called plot). 

\begin{figure}[!tbp]
\subfigure[option \& plot attention]{
\begin{minipage}[b]{0.21\textwidth}
\flushleft
\includegraphics[scale=0.39]{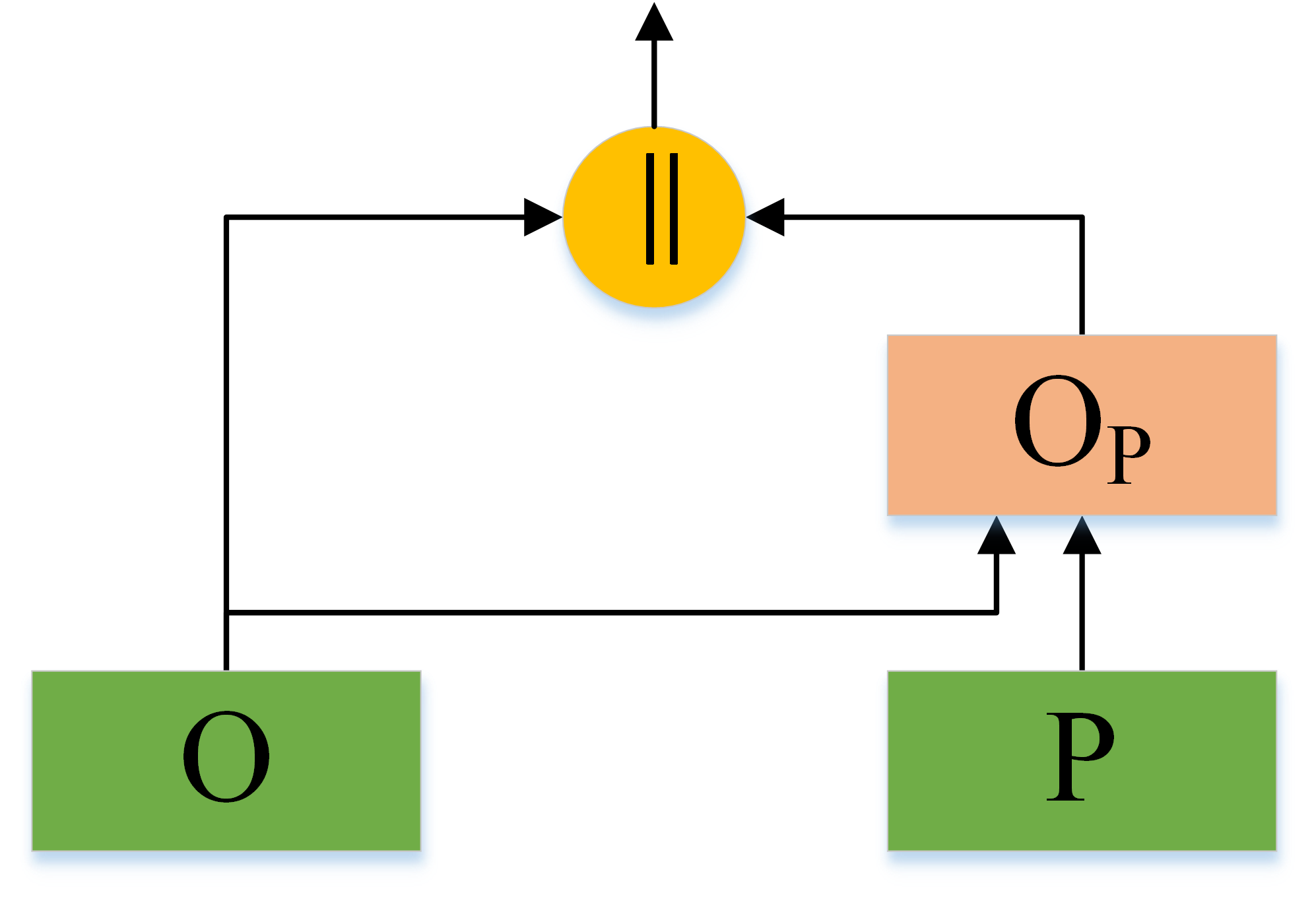}
\label{fig:pic1}
\end{minipage}
}
\subfigure[option \& climax]{
\begin{minipage}[b]{0.2\textwidth}
\flushleft
\includegraphics[scale=0.39]{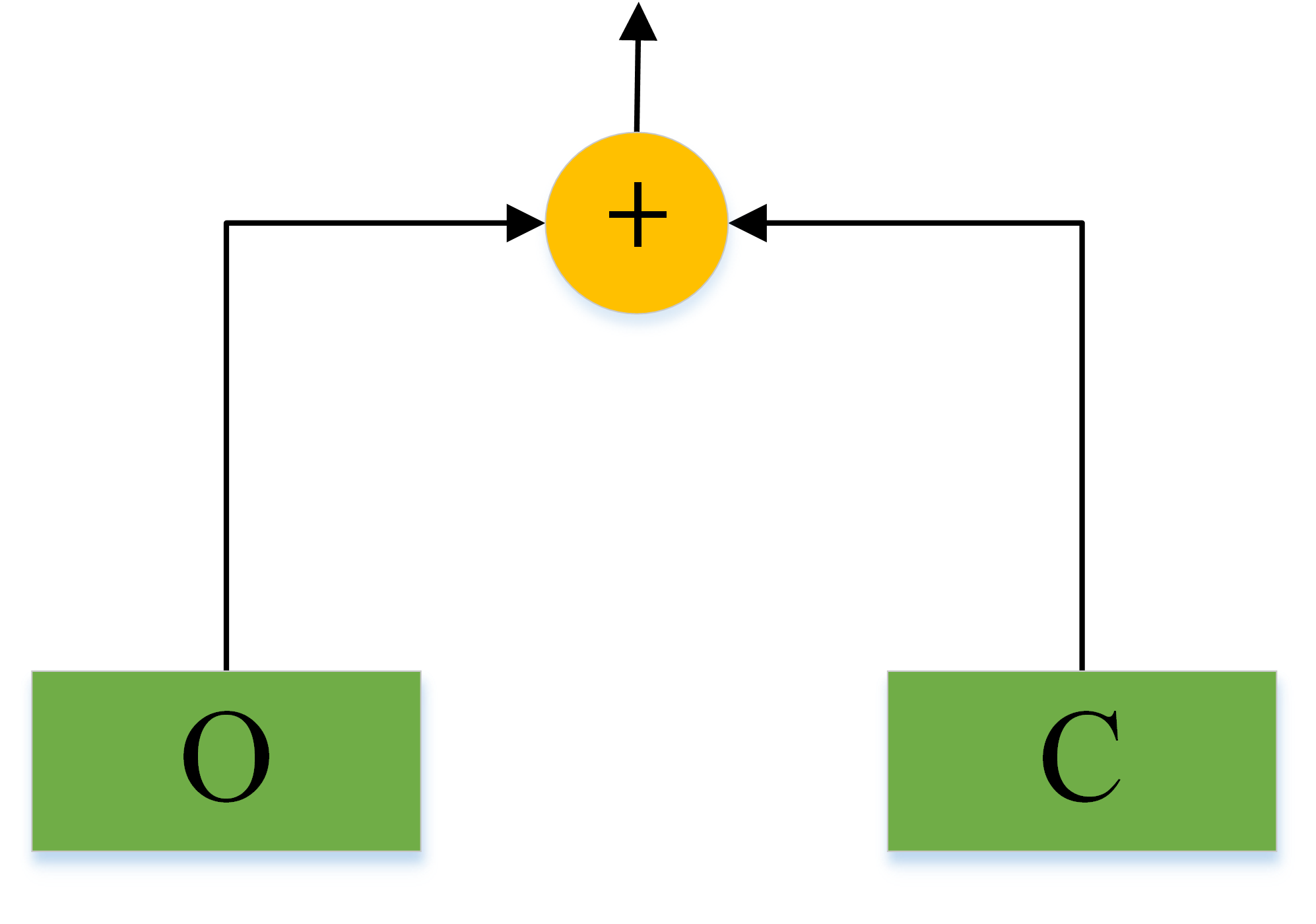}
\vspace{-0.5cm}
\label{fig:pic2}
\end{minipage}
}

\subfigure[option \& climax attention]{
\begin{minipage}[b]{0.21\textwidth}
\flushleft
\includegraphics[scale=0.39]{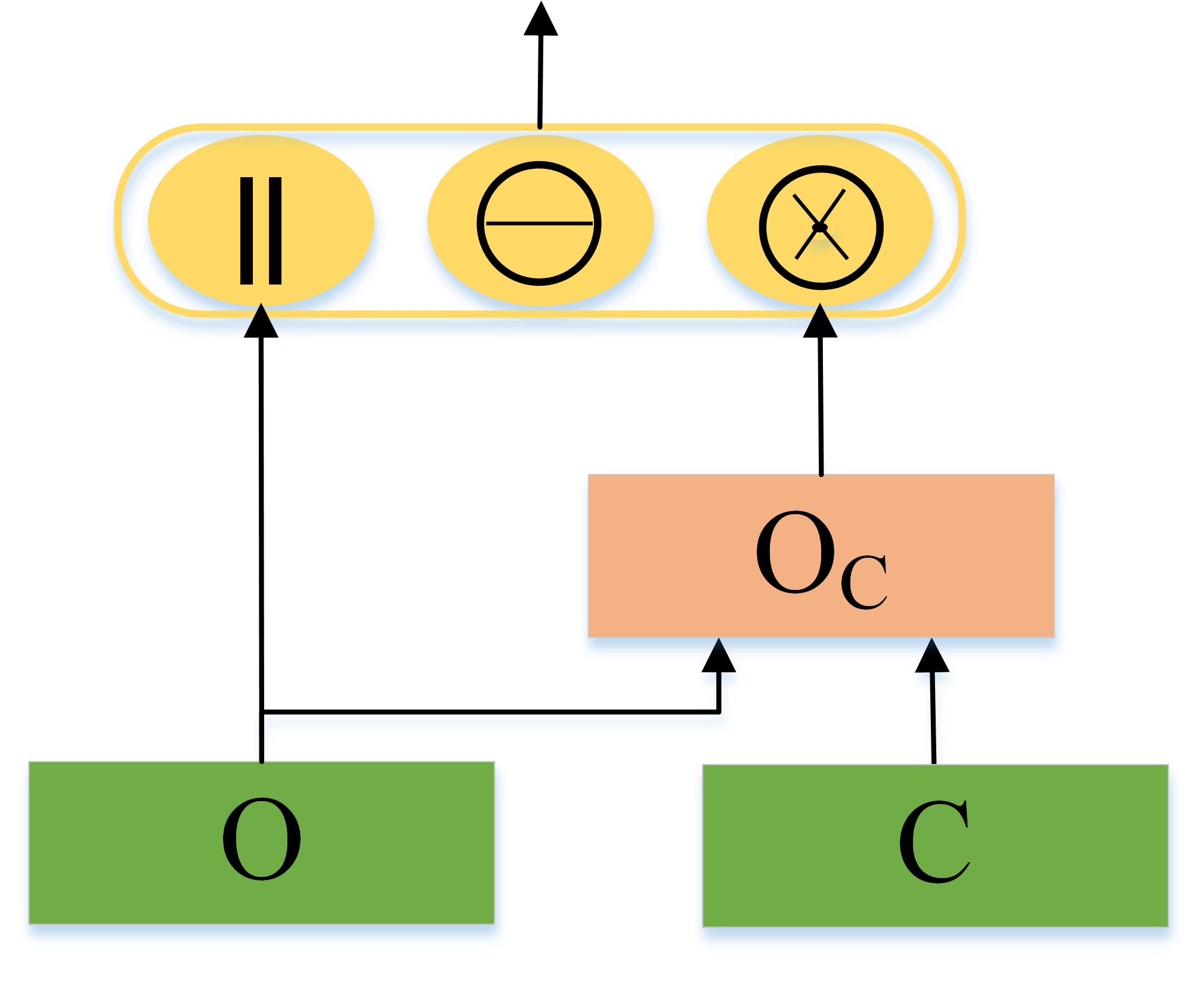}
\vspace{-0.2cm}
\label{fig:pic3}
\end{minipage}
}
\subfigure[expostion enhanced option \& climax attention]{
\begin{minipage}[b]{0.2\textwidth}
\flushright
\includegraphics[scale=0.13]{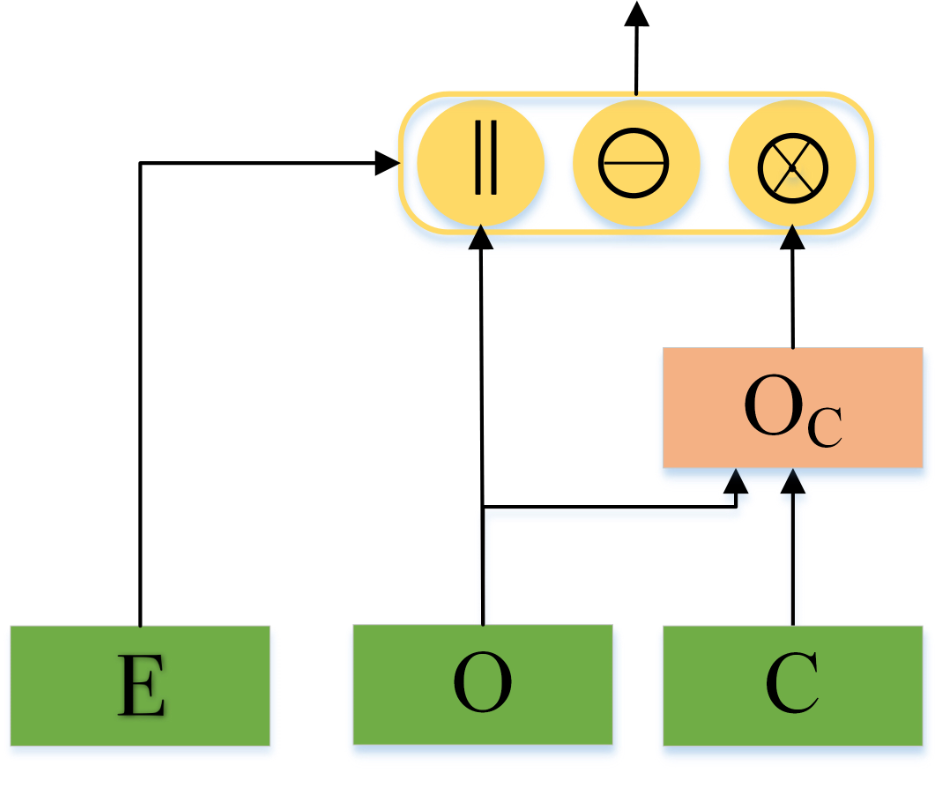}

\vspace{-0.2cm}
\label{fig:pic4}
\end{minipage}

}
\caption{\label{fig:small} Four strategies of modeling a story}
\end{figure}

Recently, methods based on linear classifier with handcrafted features, as well as neural network (NN), have been proposed for the story-cloze test.
Handcrafted features based methods \cite{Chaturvedi2017StoryCF,Mostafazadeh2016ACA,Schwartz2017TheEO} extract commonsense knowledge like events sequence and sentiments trajectory by external tools to help the story understanding.   
first Among NN based methods, the val-LS-skip model \cite{Srinivasan2018ASA} represents the last sentence in the plot and the option ending with skip-embeddings, and then processes them with a simple feed-forward neural network. Their experiments show that representing the whole four-sentence plot with the ending performs worse than only using the fourth sentence with the ending. 
\deleted{the HIER\&ATT model \cite{Cai2017PayAT} presents a hierarchical bidirectional LSTM (BiLSTM) \cite{Hochreiter:1997:LSM:1246443.1246450} with attention \cite{Bahdanau:iclr2015-nmt} to model the option ending and the plot.}

However, intuitively speaking, a coherent ending should be related to the whole plot, instead of just the fourth sentence. 
This intuition is opposite to the conclusion of \newcite{Srinivasan2018ASA}. 
To explore whether the content in a plot, except for only the fourth sentence, can assist in choosing a correct ending, 
we observed a large number of stories and discovered two phenomena: \textit{(1) The ending is usually directly affected by the last sentence in a plot. (2) The first three sentences in a plot usually provide background settings about the character, time, and place of the story, which influences the ending implicitly but essentially.} 
Inspired by our findings, we divide a complete story into three parts: an \textit{exposition part} (the first three sentences), a \textit{climax part} (the fourth sentence), and an \textit{ending}.  
Within a story, exposition is the beginning of the narrative arc. It introduces key scenarios, themes, characters, or settings about a story, and creates the rising action of the story which then reaches the climax and continues through into the resolution. In addition, we still denote the four sentences as a plot. Table~\ref{table:example_roc} shows an example from the ROCStories Corpus, consisting of the three segments we described above.
\deleted{, and one of the ending as an option.} 


Based on the narrative segments of a story, we summarize four strategies for the story-cloze test task. We show the four strategies in Figure~\ref{fig:small}, and here we denote the ending as the option. The strategy (a)  \newcite{Cai2017PayAT} treats the exposition part and the climax as a whole. The strategy (b) just considers the representations of the climax and the ending without interaction \cite{Srinivasan2018ASA}. The strategy (c) interacts the ending with the climax in word level to acquire more sufficient information. The strategy (d) uses a distilled exposition to enhance the interaction between the climax and the ending.  
Previous studies highlight the importance of the climax but ignore the importance of the exposition. How to exploit useful information from the exposition, and further help the model to understand the story is the key problem we are trying to figure out in this work. 

\deleted{(also called a climax \newcite{Chaturvedi2017StoryCF}) in a plot. This finding is consistent with the observation of \citep{Srinivasan2018ASA}. They find that considering just the last sentence instead of the whole plot performs better. Figure~\ref{fig:small}(b) shows a diagram of their model
They use skip-thought embeddings followed by a feed-forward network the climax and the ending
}

\deleted{
It is obvious that the correct ending of the story in Table~\ref{table:example_roc}  follows the climax directly.  
A more interesting discovery is that if we replace the original climax ``She didn't like how different everything was." by ``She was curious about everything new.", then the incorrect ending ``Sarah then decided to move to Europe." becomes a reasonable and correct ending.  
The first three sentences provide background about the story which influences the ending implicitly. Previous studies highlight the importance of the climax but ignored this implicit information. How to exploit useful information from the partial context to help understand the story and identify the correct ending is still a gap (???). 
}

\begin{table}
\begin{tabular}{p{0.9\columnwidth}}
\toprule
\textbf{Exposition:} Tom was studying for the big test. He then fell asleep do to boredom. He slept for five hours.   \\   
\textbf{Climax:} He woke up shocked. \\
\textbf{False-Option:} Tom felt prepared for the test.      \\
\textbf{True-Option:} Tom hurried to study as much as possible before the test. \\

\bottomrule
\end{tabular}  
\setlength{\belowcaptionskip}{-0.3cm}
\caption{\label{table:example_roc} An example from the ROCStories Corpus.}
\end{table}
\deleted{
\begin{table}
\begin{tabular}{p{0.9\columnwidth}}
\toprule
\textbf{Exposition:} John was writing lyrics for his new album. He started experiencing writer's block. He tried to force himself to write but it wouldn't do anything.   \\   
\textbf{Climax:} He took a walk, hung out with some friends, and looked at nature. \\
\textbf{False-O:}John then got an idea for his painting.      \\
\textbf{True-O:} He felt inspiration and then went back home to write. \\

\bottomrule
\end{tabular}  
\setlength{\belowcaptionskip}{-0.3cm}
\caption{\label{table:example_roc} An example from the ROCStories Corpus.}
\end{table}
}

Following the strategy (c) and (d), we propose a Distilled-Exposition Enhanced Matching Network (DEMN) for the story-cloze test. The model comprises three modules: an input module, a matching module, and a distillation module. The input module is constructed by an embedding layer with various embeddings and an encoding layer with a BiLSTM. 
The matching module matches the option ending with the climax, using a matching network proposed by \cite{chunhualiu-mimn:nlpcc2018}. And the distillation module focuses on distilling the exposition and injects it to the matching network. 
Our model does not only match the climax with the ending explicitly but also takes full advantage of the exposition. 


We conduct experiments on the ROCStories Corpus \cite{Mostafazadeh2016ACA}. Our model achieves an accuracy of 80.1\% on the story-cloze test, outperforming all previous methods. Our key contributions are as follows: 
\begin{itemize}
\setlength{\itemsep}{0pt}
\setlength{\parsep}{0pt}
\setlength{\parskip}{0pt}
\item We divide a story into three narrative segments: an exposition, a climax, and an ending. 
\item We use a matching network to model the interaction between the climax and ending explicitly.
\item We distill the crucial parts in the exposition to help identify a coherent ending, which is proved to be significantly effective.
\deleted{\item We empirically demonstrate that our model outperforms the current state-of-the-art models.}
\end{itemize}

\section{Model}
\label{sec:model}


\subsection{Model Overview}

\label{subsec:model-overview}
\begin{figure*}[!tbp]
\renewcommand{\arraystretch}{0.8}
    \centering
 \includegraphics[scale=0.4]{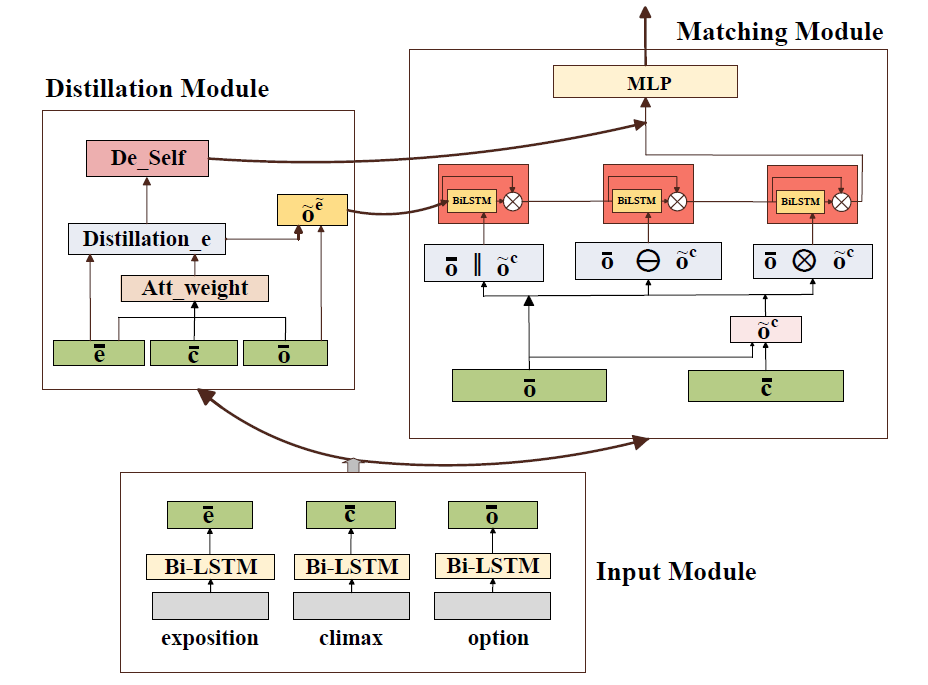}
  \vspace{-0.3cm}
  \setlength{\belowcaptionskip}{-0.5cm}
    \caption{Overview of our DEMN model.}
    \label{fig:model}
\end{figure*}

The overview of our DEMN model is shown in Figure~\ref{fig:model}. We denote $e=\{e_1,e_2,\cdots,e_{|e|}\}$ as the exposition, $c=\{c_1,c_2,\cdots,c_{|c|}\}$ as the climax and $o=\{o_1,o_2,\cdots,o_{|o|}\}$ as one of the option endings. 

\noindent\textbf{Input Module: Embedding Layer} $\quad$ This layer aims to map each word in $e, c, o$ to a semantically rich d-dimensional embedding. Following \cite{Chen2017:DrQA}, we use various embeddings to construct the d-dimensional embedding, including the pre-trained 300-dimensional Glove word embedding, the part-of-speech (POS) embedding, named entity recognition (NER) embedding, term frequency (TF) feature and exact-match feature. Furthermore, we use the relation embedding (Rel). For each word, the relationship with any other word with another sequence will be recorded.

\deleted{
For each word in a sequence $E$, if it satisfies any relation with another word in the $c$ or $o$, the corresponding relation will be taken out. The relations is extracted from ConceptNet.}

\noindent\textbf{Input Module: Encoding Layer} $\quad$ The goal of this layer is to acquire the context representation of the exposition, the climax, and the option. A single-layer bidirectional LSTM (BiLSTM)~\cite{Hochreiter:1997:LSM:1246443.1246450} is applied to transform their word embeddings respectively. Then, we can obtain the exposition hidden outputs $\bar{e} = \text{BiLSTM}(e) \in \mathbb{R}^{|e| \times 2d}$, the climax hidden outputs $\bar{c} = \text{BiLSTM}(c) \in \mathbb{R}^{|c| \times2d}$, and the option hidden outputs $\bar{o} = \text{BiLSTM}(o) \in \mathbb{R}^{|o| \times 2d}$.

\deleted{
\begin{align}
&\bar{e}_k = \text{BiLSTM}(e, k) , & i \in [1,2,  \cdots,|e|]  \label{formula:e_k} \\
&\bar{c}_j = \text{BiLSTM}(c, j) , & k \in [1,2,  \cdots,|c|]  \label{formula:c_j} \\
&\bar{o}_i = \text{BiLSTM}(o, i) , & j \in [1,2,  \cdots,|o|] \label{formula:o_i} 
\end{align}
}
\noindent\textbf{Matching Module} \deleted{: Matching Option With Climax} $\quad$ This module is responsible for matching the option $\bar{o}$ with the climax $\bar{c}$.  
It first computes word level attention vectors between the $\bar{c} $ and the $\bar{o}$. Then the attention vectors, along with the hidden outputs from the encoding layer, are matched at word-level and are further aggregated in multi-turn with the memory component. Finally, max pooling and average pooling are applied on the aggregation outputs to form a fixed length aggregation vector for the output layer. 

\noindent\textbf{Distillation Module} \deleted{Distilled-Exposition Enhancement} $\quad$ The purpose of this module is to distill the exposition and infuse it to the matching process. 
We first present how to distill the crucial information in the exposition, and then inject the information in two different fashions to enhance the whole matching process.

\noindent\textbf{Output Layer} $\quad$ This layer is used to predict which candidate ending is more reasonable. The input of this layer is the aggregation vector from the matching process. We apply a two-layer FNN with $tanh$ activations as a score function to produce the final prediction label.  

\subsection{Matching Module}
\label{subsec:matching-ec}
In this layer, we first use word-level attention to model interactions among the climax $\bar{c}$ and the ending $\bar{o}$. Then we use the multi-turn matching mechanism to compare the interactive representations $ \tilde{o}^c$ with the original hidden outputs $\bar{o}$ \cite{chunhualiu-mimn:nlpcc2018}. 

\noindent\textbf{Sequence attention} $\quad$  Here, we adopt dot product attention to model the interaction between two sequences. Given two d-dimensional vector sequences $x=\{x_1, x_2,\cdots,x_{|x|}\}$  and $y =\{y_1, y_2,\cdots, y_{|y|}\}$ with length $|x|$ and $|y|$ respectively, we define a sequence attention function $Attn(x , y) $ to compute the x-aware y representation as follows: 
\begin{align}
& Attn(x,y) = \{ {\beta}_{i}^T y \}_{i=1}^{|x|} \\
& {\beta}_{i} = Softmax(x_i\, y^T )
\end{align}
where the $\beta_i \in \mathbb{R}^{ 1 \times |y| } $ indicates how $x_i$ is relevant to each element of $y$.  

\noindent\textbf{Option-aware climax} $\quad$
\deleted{The climax carries the important information that indicates the trend of the next episode in the story. A coherent ending should follow the climax. }
For each embedding $\bar{o}_i$, to find the related parts in the climax, we compare it with the hidden outputs $\bar{c}$ to obtain the option-aware climax representation $\tilde{o}^c$, where $\tilde{o}^c = Attn(\bar{o} , \bar{c})$. 
\deleted{
\begin{align}
&\tilde{o}_i= \sum_{j=1}^{|c|}{\beta}_{ij} \bar{c}_j, &{\beta}_{ij}= \frac{exp({\bar{o}_i^T \bar{c}_j)}} {\sum\limits{_{j^\prime=1}^{|c|}exp(\bar{o}_i^T \bar{c}_{j^\prime})}} \label{formula:beta}, 
\end{align}}
For each word in the option, the relevant content in the climax will be selected and fused into $\tilde{o}_i^c$.


\noindent\textbf{Matching option with climax} \quad In order to better infer whether the option ending is semantically consistent with the climax, we use the following three matching functions to compare the $\bar{o}$ and the $\tilde{o}^c$ \cite{wang-jiang:ICLR2017,Wang:Co-matching:ACL2018} : 
\begin{align}
&u^1  = \text{ReLu}(W^1 (\bar{o} \, || \, \tilde{o}^c  ) +b^1) \label{formula:f^c}\\
&u^2  = \text{ReLu}(W^2 (\bar{o} \, \ominus \, \tilde{o}^c ) +b^2)\label{formula:f^s}\\
&u^3  = \text{ReLu}(W^3 (\bar{o} \, \otimes \, \tilde{o}^c ) +b^3) \label{formula:f^m}
\end{align}
where $||$, $\ominus$, and $\otimes$ represent the concatenation, element-wise subtraction, and element-wise multiplication between two matrices respectively. These operations can match the ending with the climax from different views.

\noindent \textbf{Multi-turn aggregation} $\quad$
In this layer we aim to integrate the matching matrices $\{u_i\}_{i=1}^{3}$ to acquire deeper understanding about the relationship between the option and the climax. Following \cite{chunhualiu-mimn:nlpcc2018}, we utilize another BiLSTM with an external memory matrix to aggregate the these matching matrices.
\begin{align}
&h^{t} = \text{BiLSTM}( W_{h} (u^{t} \,||\,  m^{(t-1)}))    \label{formula:BiLSTM} \\
&m^{t}  =   g^{t} \otimes h^{t} + (1-g^{t})\otimes m^{(t-1)} \label{formula:mem_gate}\\
& g^{t} = \sigma( W_g(h^{t} \,||\, m^{(t-1)}) + b_g) 
\end{align}
where  $W_{h}$, $W_g$, and $b_g$ are parameters to be learned, $\sigma$ is a sigmoid function, 
and $m^{(t-1)}$ is a memory vector that stores the history aggregation information. 

The last memory matrix $m^3$ stores the whole matching and aggregation of information. To obtain a global aggregation representation, we convert it to a fixed length vector with max and average pooling. The final aggregation vector $\hat{o} = MaxPooling(m^3)\,||\,AvePooling(m^3)$ . 



\deleted{\noindent \textbf{Pooling} $\quad$ We perform both max pooling and average pooling on the output of matching layer\deleted{$\{m^3_i\}_{i=1}^{l_e}$}. Then we concatenate the two pooling vectors to form a fixed-length vector \deleted{$r$} and feed it to the prediction layer.}
\deleted{
\begin{align}
&r = [ max(\{m^3_i\}_{i=1}^{l_e});avg(\{m^3_i\}_{i=1}^{l_e})].
\end{align}}

\subsection{Distillation Module}
\label{subsec:hier-exposition}
In this subsection, we focus on distilling the exposition and incorporating the distilled exposition into the matching process. 

\noindent \textbf{Exposition distillation} \quad In a four-sentence plot, the exposition generally contains abundant background information of a story. Along with the development of a story, part of the background information may become useless and noisy. To avoid the negative effect caused by redundant content, this module aims to distill the exposition to maintain the vital content and filter out the irrelevant content. The distillation process can be divided into two steps. 

The first step chooses the relevant context of the climax and the ending, \deleted{according to the exposition} by computing attention with the exposition. We compute the exposition-aware climax by $\tilde{e}^c= Attn(\bar{e}, \bar{c} )$. Similarly, we get the 
exposition-aware ending $\tilde{e}^o= Attn (\bar{e}, \bar{o})$.


The second step distills the the exposition using a carefully designed attention weight. 
\begin{align}
 &s =  (\bar{e}  \ominus \tilde{e}^c) \otimes (\bar{e} \ominus \tilde{e}^o) \\
&\alpha = softmax(s^T s) \\
&\tilde{e} = \alpha \bar{e}
\end{align}
The attention weight $\alpha$ embodies the climax and option which are carefully selected. The $\tilde{e}$ is called the distilled exposition, which is obtained by distilling the exposition, the climax, and the option. The distillation process experienced multiple information selection. In this way, the crucial parts in the exposition that related to the climax and the option can be highlighted.

\noindent \textbf{DEEM: Distilled-Exposition Enhanced Memory}\quad The distilled exposition $\tilde{e}$ organizes the relevant information about the climax and the ending, which can be used to enhance the matching process between the climax and the ending.  To make the background information flow through each turn, we infuse  the refined exposition to the initial memory component. 
We first compute the option-aware distilled exposition $\tilde{o}^{\tilde{e}} = Attn (\bar{o}, \tilde{e})$, then we use the $\tilde{o}^{\tilde{e}}$ to initialize the matching memory described in multi-turn aggregation: 
\begin{align}
m^0 = \tilde{o}^{\tilde{e}}  
\end{align}

\noindent \textbf{DEEAV: Distilled-Exposition Enhanced Aggregation Vector} \quad Word-level attended exposition can capture the relevant information in a particular view of the other word. However this lacks an overall representation about the distilled exposition itself. Summarizing all the exposition with different weights can provide the whole picture about the background settings. Hence, we first transform the exposition to a fixed-length vector with self-attention \cite{Yang:2016HierarchicalAN}. 
\begin{align}
& \hat{e}=\sum_{k=1}^{|e|} \alpha \tilde{e}, \quad & \alpha_{i} = Softmax(W \tilde{e} )
\end{align}
where the $\hat{e} \in \mathbb{R}^{2d} $ is a distilled exposition vector that summarizes all the information of the exposition according to different important degrees. 

Then, we combine the $\hat{e}$ with the output of the matching module together for final prediction. 
\begin{align}
v=  \hat{o} \,||\,\hat{e} 
\end{align}
where $v$ does not only contain the interactive representation between the climax, but also holds the whole picture about the exposition. When the distilled-exposition vector $\hat{e}$ is used, the vector $v$ is fed into the output layer to compute the probability of being a coherent option ending.



\section{Experiments}
\subsection{Experimental Settings}

\noindent \textbf{Dataset} \quad
\deleted{To verify the effectiveness of our model, w}We employ the ROCStories Corpus released by~\newcite{Mostafazadeh2016ACA} to evaluate our model. There are 100k stories in the training set and 1871 stories in both validation set and test set.
The training set contains stories with a plot and a correct option ending. The validation set and the test set both contain a plot and two option endings, including a correct one and an incorrect one. 

Following previous studies, we train only on the validation set and evaluate on the test set. During training, we hold out 1/10 of the validation set to fine tune parameters, and save the best performing parameters for testing.
For data preprocessing, we use spaCy \footnote{https://github.com/explosion/spaCy} for sentence tokenization, Part-of-Speech tagging, and Name Entity Reorganization. The relations between two words are generated by ConceptNet \footnote{http://conceptnet.io/}. 

\vspace{1mm}
\noindent\textbf{Model Configuration} \quad
\begin{table}[!tbp]
\centering
\begin{tabular}{l|c}
\toprule
Models & Test-Acc\\
\hline
DSSM \cite{Mostafazadeh2016ACA} &58.5\%  \\
HIER\&ATT \cite{Cai2017PayAT}  &74.7 \% \\ 
MSAP \cite{Schwartz2017TheEO} &75.2\% \\
val-LS-skip \cite{Srinivasan2018ASA}  &76.5\% \\
HCM \cite{Chaturvedi2017StoryCF}&77.6\% \\
Memory Chains \cite{Liu:acl2018} & 78.5\% \\
\hline
\textbf{option-climax } (single) & \textbf{77.8\%} \\ 
\textbf{DEMN} (single) & \textbf{80.1\%} \\
\textbf{DEMN} (ensemble) & \textbf{81.2\%} \\
\bottomrule
\end{tabular}
\caption{\label{table:hemn-result}Test accuracy of the SOA models.
\deleted{Our DEMN outperforms competitive baselines and state-of-the-art model.}  }
\end{table}

We implement our model with Pytorch \footnote{http://pytorch.org/}. We initialize the word embeddings by the pre-trained 300D GloVe 840B embeddings \cite{Pennington:emnlp2014} and keep them fixed during training. 
The word embeddings of the out-of-vocabulary words are randomly initialized. 
 We use Adam \cite{Kingma:ICLR2014} for optimization. 
As for hyper-parameters, we set the batch size as 64, the learning rate as 0.008, the dimension of BiLSTM and the hidden layer of MLP as 96, the L2 regularization weight decay coefficient as 3e-8, the dropout rate for word embedding and the initial memory in multi-turn aggregation as 0.4 and 0.41 respectively. The dimension of POS embedding, NER embedding, and Rel embedding are set as 18, 8, and 10 respectively. They all are fine-tuned during the training.  

\subsection{Results and Analysis}



\noindent\textbf{Baselines} \quad Table~\ref{table:hemn-result} shows the results of our DEMN model along with the published models on this dataset. The DSSM model is reported in \newcite{Mostafazadeh2016ACA}, which applies an LSTM to transform the four-sentence plot and the option to corresponding sentence vectors. Based on these vectors, the model chooses the option that has higher cosine similarity with the plot. Both the HIER\&ATT and MSAP show that using ending alone can get better performance.  The HIER\&ATT model uses hierarchical BiLSTM with attention to encode the plot and the ending. The MSAP model trains a linear classifier model with stylistic features and language model predictions. The val-LS-skip model simply sums the skip-embeddings of the climax and the ending for final prediction. Both the HCM model and the Memory Chains model exploit three aspects of semantic knowledge, including the event sequence, sentiment trajectory, and topic consistency to model the plot and the ending. However, they use these features differently. The HCM model designs hidden variables to weight the three aspects, while the Memory Chains model leverages a neural network with memory chains to learn representation for each aspect. 

\noindent\textbf{DEMN Results Analysis} \quad
The option-climax model only uses the climax to match with the option. This model reaches an accuracy of 77.8\%, which provides a strong baseline for infusing the exposition into the interaction process.  
Our DEMN model achieves an accuracy of 80.1\%, outperforming the current state-of-the-art result \cite{Liu:acl2018} with 1.6\% absolute improvement. 
Our model exceeds the HIER\&ATT by 5.4\% in terms of accuracy. 
We attribute this improvement to separately handling the exposition and the climax, and the multiple word-level attentions. 
Comparing with the val-LS-skip model, our DEMN model yields 3.6\% improvement, which proves that the exposition can promote the model effectively instead of demoting. This conclusion is exactly opposite to \newcite{Srinivasan2018ASA}. 


\subsection{Ablation Study}
\begin{table}[!tbp]
\centering
\begin{tabular}{p{0.2\columnwidth}|l|cc}
\toprule
distillation & Models & Test-Acc\\
\hline
& \textbf{Full Model} & 80.1\% \\
distillation & w/o exposition-mem & 78.5\% \\
exposition & w/o exposition-vec &78.8\% \\
\hline
w/o& exposition-mem\&vec & 79.3\% \\
distillation  & w/o exposition-mem & 78.8\% \\
exposition& w/o exposition-vec &78.6\% \\

\bottomrule
\end{tabular}
\caption{\label{table:exp-distillation}Ablations studies about the influence of distilling exposition.
}
\end{table}

To evaluate how different components contribute to the model performance, we design two groups of experiments. We will discuss the influence of distilling the exposition and different ways of distillation. 

\noindent\textbf{Influence of distilling exposition} \quad We conduct ablation study on two different ways of using the distilled exposition, by removing the distilled-exposition memory and distilled-exposition vector. Furthermore, to verify the effectiveness of the distillation, we design another three experiments without the distillation process. In this case, we use the original BILSTM hidden outputs of exposition $\bar{e}$ to replace the distilled exposition $\tilde{e}$. Table~\ref{table:exp-distillation} shows the experimental results.

The first group of results in Table~\ref{table:exp-distillation} shows the performance of removing two kinds of exposition separately. We observe a substantial drop, when we replace the distilled-exposition memory with zero memory. For each word in the option, the corresponding distilled-exposition memory can provide the different background knowledge about the story. This result proves that the information in the exposition is vital for choosing a correct option. On the other hand, removing the distilled-exposition vector impairs the performance as well.
Compared with removing exposition memory, the accuracy declines less when removing exposition vector, indicating the former offers more benefits to the model.

The second group of results in Table~\ref{table:exp-distillation} report the performance without distillation. We observed that the highest accuracy 79.3\% appears when two kinds of exposition representation are incorporated into the model. Once again, these results show that the content in the exposition is helpful in choosing a coherent option-ending. 
We can also see that the models would be impaired, no matter which component is removed. However, the accuracy drops slightly than the first group (1.6\% absolute vs 0.5\% absolute, and 1.3\% absolute vs 0.7\% absolute). 

\begin{table}[!tbp]
\centering
\begin{tabular}{l|c}
\toprule
Models & Test-Acc\\
\hline
\textbf{Full Model} & 80.1\% \\
\hline
w/o exposition-aware climax & 79.7\% \\
w/o exposition-aware option &78.9\% \\
w/o exposition-aware climax\&option  &78.6 \% \\ 

\bottomrule
\end{tabular}
\caption{\label{table:weight-exposition} Ablation studies about different ways of distilling exposition.
}
\end{table}

\begin{figure*}[!htbp]
\subfigure[memory1]{
\begin{minipage}[b]{0.32\textwidth}
\vspace{0pt}
\includegraphics[scale=0.25]{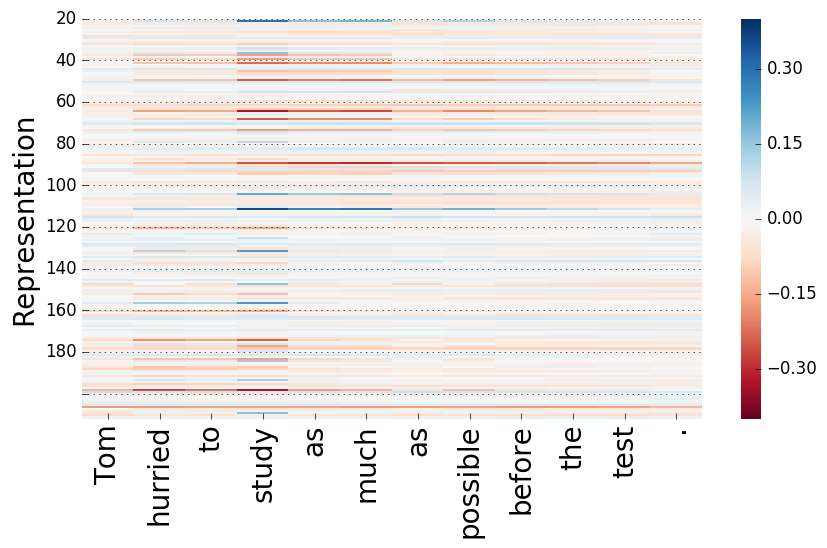}
\label{fig:pic2}
\end{minipage}
}
\subfigure[memory2]{
\begin{minipage}[b]{0.32\textwidth}
\includegraphics[scale=0.25]{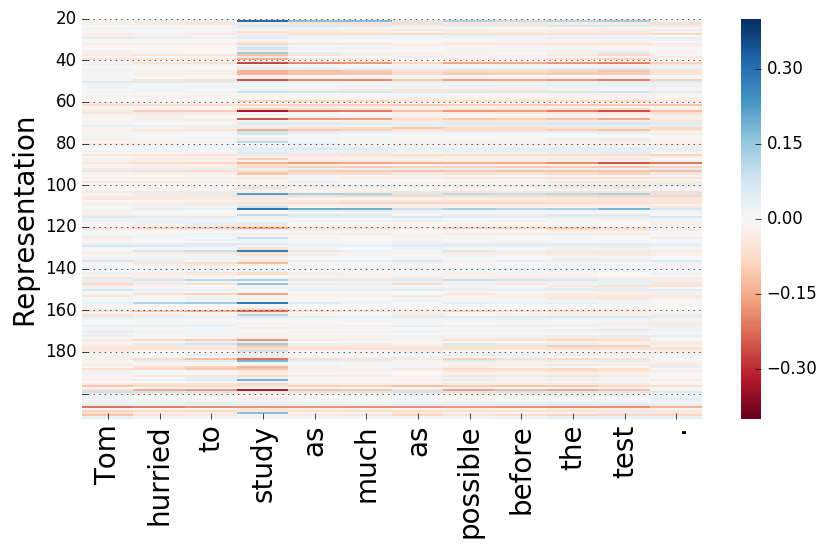}
\label{fig:pic3}
\end{minipage}
}
\subfigure[memory3]{
\begin{minipage}[b]{0.32\textwidth}
\includegraphics[scale=0.25]{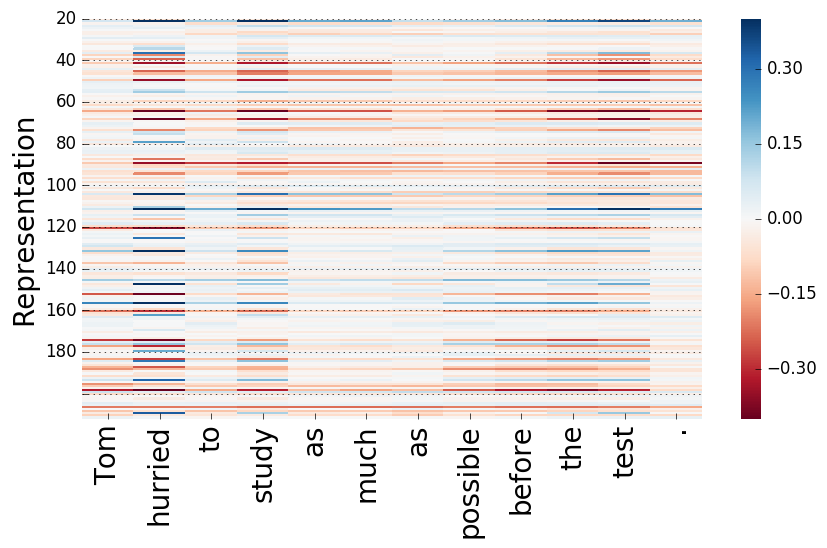}
\label{fig:pic4}
\end{minipage}

}
\caption{\label{fig:mems} The memory representations in three turns}
\end{figure*}

\noindent\textbf{Different ways of distilling exposition} \quad
To investigate the influence of the attention weight to distill the exposition, we conduct an ablation study on the way of calculating the attention scores. In our DEMN model, the attention weights imply three aspects of information, including exposition itself, the exposition-aware climax, and the exposition-aware option. We observe that the accuracy is slightly influenced without the exposition-aware climax. While removing the exposition-aware option affects the performance more obviously. 
The degrees of performance degradation are various. Removing both of the interactive representations is the most detrimental. These reflect that distilling more accurate exposition is important.

\deleted{
\begin{table}[!htbp]
\begin{center}
\begin{adjustbox}{max width=\linewidth}
\begin{tabular}{l|cc}
\toprule
Models & Multi & Single  \\
\hline
\text {C } &  79.4\%  &79.6\%\\ 
\text {S } & 77.3\%  & 78.7\%\\
\text {M}& 74.9\% &  54.9\% \\
\text {CS} & 78.6\% &79.8\%\\
\text {CM} &79.3\%  &79.6\%\\
\text {SM} &78.7\%  &77.7\% \\
\text {CSM} & \bf{80.1\%}  &79.3\%\\
\bottomrule
\end{tabular}
\end{adjustbox}
\setlength{\belowcaptionskip}{-0.5cm}
\caption{ \label{table:result of multi vs single} Test accuracy of Multi-turn and Singl-turn.}
\end{center}
\end{table}
}

\subsection{Discussion}
In this work, we adopt three matching features to exploit the relation between two sequences in word level, which plays an important role in matching network.  
To discuss how these matching features influence the model, we design a series of experiments and give further analysis on the learning curves of the test set. 
\begin{figure}[!ht]
    \centering
  \includegraphics[width=0.45\textwidth]{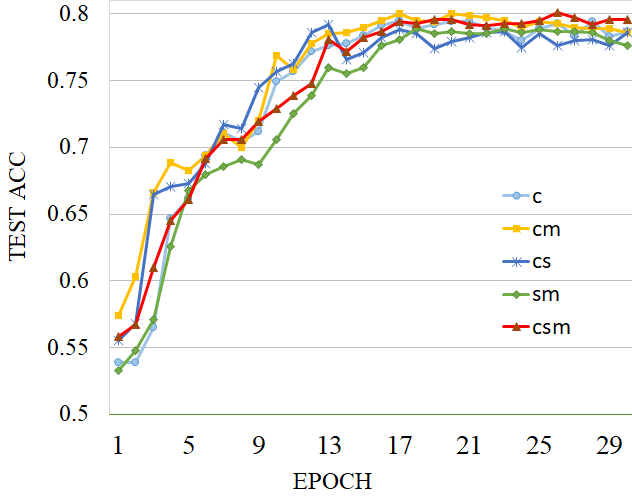}
    \caption{Learning curves of different matching features}
    \label{fig:test-accuracy-trend-m}
\end{figure}

\deleted{Figure~\ref{fig:test-accuracy-trend-s} show learning curves of using one single matching feature. We can see that the feature obtained from concatenation is the best performing, while the subtraction feature is inferior to the concatenation. Among the three methods, the multiplication features get the lowest accuracy. These results are consistent with our intuition since both the subtraction and multiplication operations cause information loss.}


Figure~\ref{fig:test-accuracy-trend-m} presents the performance of using different matching features. Both curves of the ``cs" and ``cm" rise obviously in the early training stage, however, the ``cs" 
suffer a sustained decrease after the peak. 
The two curves of ``cm" and ``csm" are very close in the latter training stage. Until the epoch 26, the ``csm" reaches an accuracy of 80.1\% on the test set with an accuracy of 81.8\% on the validation set. We finally choose the model with ``csm", hence it performs better on the validation set.     

\subsection{Visualization}
To gain an insight into the multi-turn aggregation process employed by the model, we observed many visualized pictures of the memory representations during the multi-turn process of different examples. There is an obvious conclusion. From the pictures of memory1 to memory2, then to memory3: the colors of the keyword parts of the pictures are more and more prominent and obvious. Take the case in Table~\ref{table:example_roc} for example. We can see that in Figure~\ref{fig:mems}, the key word ``study'' is the captured in the memory1 and held its importance in the next two turns. In the memory2, we can see that ``the test" is marked. As both ``study'' and ``the test'' are directly relevant to the topic described in the exposition. The most surprising thing is that the model pays attention to the word ``hurried'' in the final turn, which can be highly linked to the rest parts of scenario. 
In a summary, the important words are highlighted step by step along the multi-turn process. 

\deleted{As shown in Figure~\ref{fig:mems}, the parts of the word “hurried”,"test" are highlighted step by step along the multi-turn process, which is helpful to judge whether the ending is correct for the plot. In the first two pictures, they focus on the wrong words and phrase, but in the last picture of the multiple turns, the keywords are clearly identified.}
\deleted{Second, by comparing the single-turn and multi-turn of c-infer-emb, after several rounds of learning, the special phrase part is clearly demarcated, and the multi-training helps to judge the correctness of the sentence, which is more prominent (focus is accurate): (The keywords in the last picture of the multiple rounds is clearly identified by, but the single-turn may focus on the wrong word or phrase)}

\begin{table*}[!tbp]
\renewcommand{\arraystretch}{0.9}
\begin{tabular}{p{2\columnwidth}l}
\toprule
\textbf{Exposition:} Collin likes to dress up. One Halloween he decided to wear his costume to the office. Collin's boss did not permit costumes to be worn in the workplace. \\
\textbf{Climax:} He received a write-up. \\
\textbf{False-Option:} Collin received a raise and a promotion. 
\quad\textbf{True-Option:} Collin was upset with his boss.\\
\toprule
\textbf{Exposition:} Amy was visiting her best friend in Phoenix. It was her first time there. She was excited to see the town.   \\
\textbf{Climax:} She exited the airport and was struck by the heat. \\
\textbf{False-Option:} Amy began shivering.   
\quad\textbf{True-Option:} Amy began to sweat.\\
\bottomrule
\end{tabular}  
\setlength{\belowcaptionskip}{-0.3cm}
\caption{\label{table:error_cases} Error cases}
\end{table*} 
\subsection{Error analysis}

In order to analyze which kind of problems can not be solved by our model, we observe some error cases. We find that it is difficult for the model to choose the right ending when the plot have lots of negation words, complicated phrases, or unrelated noisy words.  Table~\ref{table:error_cases} shows two error cases selected randomly. 
In the first case, we observe that the ending cannot be chose correctly only based on the climax. Taking the whole plot into consideration, the correct is obvious to us. However, because the model misunderstands the advanced phrase ``write-up" and unable to capture sufficient information from the exposition, the score of false option is much higher than the true option. 
In the second case, the two endings cover too many same words and their prediction scores are quite close. We observe that the word ``heat" is the most crucial factor of the right choice and it occurs only once. The model was puzzled by other misleading words like ``excited". So it is very challenging for the model to fully filter all the redundant information.


\section{Related Work}

The Story Cloze Test \cite{Mostafazadeh2016ACA} is proposed to evaluate the story understanding ability of a system. Recently, several neural networks are used to tackle this task. 
\textbf{HIER\&ATT} \cite{Cai2017PayAT} employs LSTM units to hierarchically encode both the ending and the ending-aware full context. A surprising finding is that only relying on the ending can achieve an accuracy of 72.7\%. \textbf{Val-LS-skip} \cite{Srinivasan2018ASA} achieves a competitive result by using a single feed-forward neural network with pre-trained skip-embeddings of the last sentence and the ending. Their experiments show that the performance of using the whole plot is worse than just using the forth sentence. However, our DEMN model can extract useful information from the exposition part, which actually improves the accuracy rather than decreases it.  

To explore the external knowledge to help the story understanding, three semantic aspects are frequently used, including events sequence, sentiment trajectory, and topical consistency \cite{Lin2017ReasoningWH,Chaturvedi2017StoryCF,Liu:acl2018}. The current state-of-the-art model \textbf{Memory Chains} \cite{Liu:acl2018} adopts the EntNet \cite{Henaff:ICLR2017:EntNet} to track the three semantic aspects of the full context with external neural memory chains.\deleted{, where each memory chain is supervised by the semantic binary label acquired from external knowledge source.}  


Another closely related work is the matching network \cite{chen-zhu:acl2017,wang-jiang:ICLR2017,chunhualiu-mimn:nlpcc2018}, which is commonly used in natural language inference \cite{MacCartney:2009}. The matching network can match the interactions between two sequences effectively. Among them, the \textbf{MIMN} model proposed by \cite{chunhualiu-mimn:nlpcc2018} introduce a multi-turn inference with memory mechanism to compute three heuristic matching \cite{Mou-EtAl:2016:P16-2} representation between two sequences iteratively. Multi-turn inference can capture more detailed content about the interactions between two inputs, and it performs well on small-scale datasets. We use the multi-turn inference method to model the interactions between the climax and the ending. Furthermore, this paper focuses on exploiting the available information in the exposition to assist the interactions of the climax and the ending. 


\section{Conclusion}

In this paper, we propose the Distilled-Exposition Enhanced Matching Network model for the story-close task. Our model achieves an accuracy of 80.1\% on ROCStories Corpus, outperforming the current state-of-the-art model.
In our task, we divide the story into an exposition, a climax, and an ending. The experimental result shows that matching the ending with the climax can achieve a strong baseline. This indicates that the interaction between the climax and the option is necessary. 
Further, we propose a method of distilling the exposition with the evidence provided by the climax and the ending. More specifically, we integrate the distilled exposition into the matching process in two ingenious manners, and yield a significant improvement. 

\section*{Acknowledgements}
This work is funded by Beijing Advanced Innovation for Language Resources of BLCU (TYR17001J), and The National Social Science Fund of China (16AYY007).

\bibliographystyle{acl}
\bibliography{paclic2018}






 \end{document}